# Cellular Automata based adaptive resampling technique for the processing of remotely sensed imagery

S.K. Katiyar                                                                P.V. Arun

Department of Civil Engineering
MA National Institute of Technology, India

**ABSTRACT**

Resampling techniques are being widely used at different stages of satellite image processing. The existing methodologies cannot perfectly recover features from a completely under sampled image and hence an intelligent adaptive resampling methodology is required. We address these issues and adopt an error metric from the available literature to define interpolation quality. We also propose a new resampling scheme that adapts itself with regard to the pixel and texture variation in the image. The proposed CNN based hybrid method has been found to perform better than the existing methods as it adapts itself with reference to the image features.

## INTRODUCTION

Intensity interpolation or resampling techniques form an integral part of different processing stages of the images and hence is important in many fields such as medical imaging, consumer electronics, military applications etc. Registration or geometric correction of images usually requires the resetting of image framework which in turn results in variation in source and target pixel sizes. When the output pixel size is not the same as the original, quality of the resampling technique determines the quality of output (Moreno, 1994). This is not only true in the visual appearance of the images, but also in the numerically interpolated values when used in multi-temporal or multi-sensor studies. In particular, rectification, registration, and geo referencing requires that an image be resampled onto a new coordinate grid. Planimetric measurements of the imagery by assignment of geographical co-ordinates are accomplished by geo referencing of images (Lillesand, 2004).

Resampling may cause the information to be lost, which is of particular importance in the case of satellite and aerial imagery. As satellite imaging systems are designed with the tradeoff between aliasing and image blurring in mind we deal with imagery whose salient features range from large low-frequency objects to sub pixel features (Scott, 2004). Remote sensing imagery is typically under sampled, i.e., we are interested in image features which may be smaller than the nominal spatial resolution of the sensor. The sub-pixel features that cannot be resolved may still be detected through local contrast and hence it is important to preserve as much of the local contrast as possible when registering such features (Storey, 2001). That is, if an object is smaller than a pixel but has a significantly different intensity compared to its surroundings, then the local contrast around that pixel ensures that the feature is distinguishable. In practice, interpolation techniques are limited by the finite extent of the image and the finite extent of the interpolating function (Kenneth et al, 2001). This generally results in a tradeoff between computational complexity and nearness to the ideal. Furthermore, when images are under sampled, the assumptions for ideal reconstruction are violated, and hence the concept of the most accurate reconstruction must be investigated.

Literature reveals a great deal of image resampling techniques and selection of an optimum image resampling method in the processing of remotely sensed data is a difficult task (Schowengerd, 1997; Parker, 1987). The arrival of new generation of satellite sensors with improved spatial and radiometric resolution has led greater demands on the resampling algorithms. In the use of remotely sensed data for the mapping of heterogeneous environment like urban areas different resampling methods perform differently in respect of preserving the image edges, which in turn are important inputs for the identification of land use features (Jordi et al., 2007). The sensor point-spread function (PSF) has also an important role in the analysis of radiometric aspects related to the resampling of remotely sensed (Nilback, 1986). Although, for the image resampling standard methods are available, but there are certain reasons, which make the selection of resampling method application specific. The first reason is that the ideal intensity interpolation function is the sinc-function, which is difficult to implement mathematically due to its infinite extent and the second is that interpolation of satellite images are not regularly sampled from a continuous grid (Stephen & Schowengerdt., 1982). In the geometric correction of remotely sensed data, the image resampling has got certain issues due to changes in the image radiometry, which affects the accuracy of subsequent image analysis such as classification and segmentation (Parker, 1987). Thus, many simpler interpolants of bounded support have been investigated in the literature.

The choice of resampling kernels depends on the intended use of the data. The simplest method of resampling is Nearest Neighbor (NN) which performs no interpolation but selects the nearest pixel value. NN resampling does not alter the brightness values of the original image. However, NN resampling is not as visually appealing as other kernels due to its blocky effects in the image (Thurman & Fienup, 2009). Bilinear (BL) is a simple 2-point linear interpolator which uses the neighboring two points to produce a smoothing effect to the image however is not used in most applications as it alters the actual image radiance (Bruce & Hilbert, 2004). Cubic Convolution (CC) is a 4-point kernel based on cubic splines and has been used for remote sensing image analysis as it provides a reasonable compromise between accuracy and speed (Nilback, 1986). The CC kernel has a slight edge enhancing effect on the images and the interpolation errors of the CC kernel are significantly worse than a 16-point kernel.

16-point Kaiser-Damped Sinc (KD16) is the most accurate as compared to the above said techniques and is based on a 16-point sinc function windowed by a Kaiser window (Moreno & Melia, 1994). Internal studies have shown that it produces a pleasing balance between image ripple and low frequency accuracy. The MTF resampling kernel is based on an empirical modeling of the optical and electronic properties of the ETM+ sensor. MTF resampling kernel is only recommended for map and ortho corrected images and may introduce a slightly blocky appearance to the more homogeneous areas of Landsat imagery (Bruce & Hilbert, 2004).

Our goal is to resample images in way by preserving the high frequency artifacts, as it is a source of the local contrast phenomena that allows the detection of sub-pixel features. This would suggest that nearest neighbor interpolation should be ideal because it does not perform pixel mixing when up sampling to a grid. However, this desired behavior is gained at the expense of assuming that the original signal contains no large scale features. If the ground truth consisted of large regions with smoothly varying intensities, cubic spline interpolation would be expected to perform much better (Oliver et al., 2006). Typical non-adaptive interpolation methods such as nearest neighbor, bilinear, and cubic resampling yield decreasing degrees of high-frequency information fidelity.

In general, higher is the order of interpolation, the smoother is the resampled image and lesser is the local contrast information (Australian Geo-Portal, 2012). Based on these discussions, we conclude that when registering an under sampled image, choosing the optimal interpolation technique requires not only knowledge of the sampling parameters, but also information about the content of the scene being imaged. Hence we propose a new interpolation scheme which dynamically takes in to account the type of feature to be resampled to facilitate its faithful reconstruction. Hilbert Schmidt Independence criteria as well as Laplace pyramid representation (Adelson & Burt, 1987) along with probabilistic rule based strategy has been used for an effective contextual representation. In this paper we adopt an intelligent hybridization of the existing efficient interpolation strategies by using the advanced random modeling techniques such as Cellular Neural Network (CNN) (Chua & Yang, 1988). Thus proposed approach maintains both the sub pixel detection capabilities and accurate interpolation of large scale structures using the scale space to adjust the resampler with reference to image features.

## DATA RESOURCES AND STUDY AREA

The investigations of present research work have been carried out for the satellite images of different spatial resolution sensors for the Bhopal city in India and details of the same are given in the table-1. The study area is new market area of Bhopal city having area central point coordinates 23° 55' N Latitude and 76° 57' E Longitude. The algorithms have been implemented in the MATLAB software environment and the accuracy of the techniques was verified using Erdas-Imagine & Matlab software. The Erdas-Imagine v9.1 was also used for the pre-processing and other analysis tasks of remote sensing satellite imagery.

Table 1: Data sources description (Source: NRSC, NASA)

| S.NO | Imaging sensor | Resolution(m) | Satellite | Area | Date of Acquisition |
|------|---------------|---------------|-----------|------|---------------------|
| 1 | LISS-III | 23.5 | IRS P5 | Bhopal(India) | 5$^{th}$ April 2009 |
| 2 | LISS-IV | 5.6 | IRS P5 | Bhopal(India) | 16$^{th}$ March 2010 |
| 3 | CARTOSAT-1 | 2.5 | IRS P5 | Bhopal(India) | 5$^{th}$ April 2009 |
| 4 | Google Earth | NA | NA | Bhopal(India) | 16$^{th}$ March 2010 |

**RESEARCH METHODS**

We propose an intelligent hybridization of the above discussed methods using computational techniques over Laplacian pyramid representation of the sampled image. The different interpolation techniques are combined at different scales as well as based on the scene features in the image. Technique behave similar to nearest neighbor while mitigating the extreme aliasing behavior typically seen with that method because it uses cubic interpolation, bilinear, KD16 at suitable scales. An important consequence of this approach is that accuracy will be enhanced at various situations, especially when an object is smaller than a single pixel but exhibits high local contrast. This is because our approach gives similar results to nearest neighbor at very fine scales, but with a cubic spline interpolant's structure superimposed. This structure comes from levels farther down in the Laplacian pyramid, where the local contrast from the immediate neighborhood at every scale is effectively combined with the sub-pixel feature.

CNN network trained using the rule inversion technique along with probabilistic rules is used to adjust the resampler according to contextual variations. Hilbert Schmidt Independence criteria are used to optimize the training by reducing the number of training samples required. Image is transformed into its Laplacian pyramid representation and high frequency information of sub-pixel objects contained in the first level of the pyramid is preserved through intelligent selection of resamplers. Our proposed approach is to perform nearest neighbor interpolation on the first level in the pyramid, while cubic spline interpolation is used for subsequent levels. The Laplacian pyramid transformation is then inverted to obtain the registered image.

**RESULTS**

The resampling accuracy of different methods has been analyzed using relevant statistical parameters mentioned in literatures (Zitova & Flusser, 2001). We have adopted error metrics as well as average difference error for the purpose. Entropy is a statistical measure of randomness that can be used to characterize the texture of the input image and has been used to measure the order of deviation of interpolated image from the original. The entropy deviations as well as the correlation values were calculated between the resampled versions of the image and original version of the image were compared to estimate the accuracy of interpolation.

We have considered the quantity $D_\alpha (j, i) = I_\alpha (j, i) - G(j,i)$ where $I_\alpha (j, i)$ is the interpolant, G is the ground truth, and $D_\alpha$ gives a simple difference error at each pixel. The ground truth value has been obtained using Google earth and was cross validated by the control points obtained by DGPS survey over MANIT campus. A lower value of difference error, and entropy deviation as well as a higher value of correlation indicates better resampling technique. Different parameters have been calculated in MATLAB and results are summarized in table 2.

The visual interpretation of the resampled images using different classical methods and proposed method on different sensor data is presented in the following figures. The difference error values and entropy values discussed in the Table 1 have shown that the intelligent CNN based approach is far better than the existing methods as it has a lesser average difference error, lesser entropy deviation and greater correlation. The performance of the proposed methodology when compared to the existing methods is evident from the results of the down sampling performed over Cartosat imagery and PAN sensor imagery presented in Figure 1 & 2. The proposed methodology is found to perform better for the up sampling performed over the LISS 3 sensor and Google earth imagery as can be noted from the Figure 3 & 4. The visual comparative analysis with reference to the terrain features over the results also reveals the optimality of our approach.

The visual results presented clearly indicate that the high fidelity features (shown in circle) are preserved well in our approach. The road in the urban area in Cartosat imagery (fig.1), the roads in the PAN imagery (fig.2), Junction in LISS 3 sensor data (fig.3), Stadium in Google Earth image (fig.4) etc. can be easily distinguished by proposed methodology when compared to classical approaches as our approach preserves the sub pixel frequencies.

Table 2: Comparative Performance Analysis of Different Resampling Techniques

| S. NO | Image Sensor | Algorithm for Resampling | Correlation Coefficient | Entropy Deviation | Average Difference Error |
|---|---|---|---|---|---|
| 1 | PAN (CARTOSAT-1) | NN | 0.84 | 0.50 | 0.62 |
| | | BL | 0.77 | 0.47 | 0.51 |
| | | CC | 0.69 | 0.44 | 0.42 |
| | | DS16 | 0.67 | 0.43 | 0.48 |
| | | KD16 | 0.68 | 0.39 | 0.32 |
| | | CNN | 0.82 | 0.21 | 0.35 |
| 2 | LISS-III | NN | 0.73 | 0.43 | 0.52 |
| | | BL | 0.69 | 0.42 | 0.48 |
| | | CC | 0.57 | 0.35 | 0.41 |
| | | DS16 | 0.54 | 0.42 | 0.32 |
| | | KD16 | 0.43 | 0.48 | 0.21 |
| | | CNN | 0.73 | 0.21 | 0.31 |
| 3 | LISS-IV | NN | 0.84 | 0.28 | 0.46 |
| | | BL | 0.71 | 0.31 | 0.21 |
| | | CC | 0.61 | 0.32 | 0.23 |
| | | DS16 | 0.73 | 0.43 | 0.19 |
| | | KD16 | 0.51 | 0.31 | 0.16 |
| | | CNN | 0.82 | 0.23 | 0.15 |
| 4 | LANDSAT-5TM | NN | 0.87 | 0.35 | 0.30 |
| | | BL | 0.77 | 0.25 | 0.23 |
| | | CC | 0.73 | 0.23 | 0.45 |
| | | DS16 | 0.73 | 0.24 | 0.10 |
| | | KD16 | 0.78 | 0.23 | 0.18 |
| | | CNN* | 0.82 | 0.21 | 0.14 |
| | | MTF | 0.87 | 0.23 | 0.09 |
| 5 | Google Earth | NN | 0.8752 | 0.312 | 0.61 |
| | | BL | 0.8542 | 0.15 | 0.32 |
| | | CC | 0.7941 | 0.24 | 0.45 |
| | | DS16 | 0.7167 | 0.261 | 0.38 |
| | | KD16 | 0.628 | 0.185 | 0.21 |
| | | CNN | 0.935 | 0.12 | 0.12 |

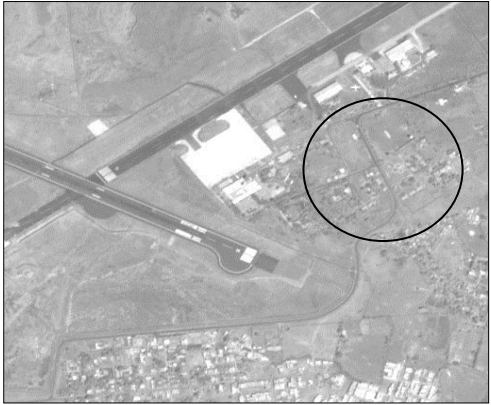
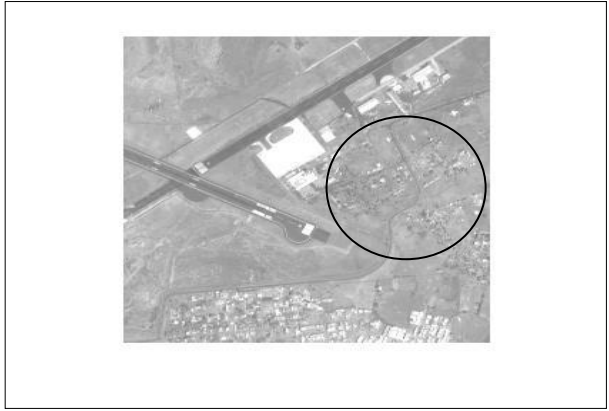

a) Cartosat imagery　　　　　　　　　　　　b) Nearest Neighbor

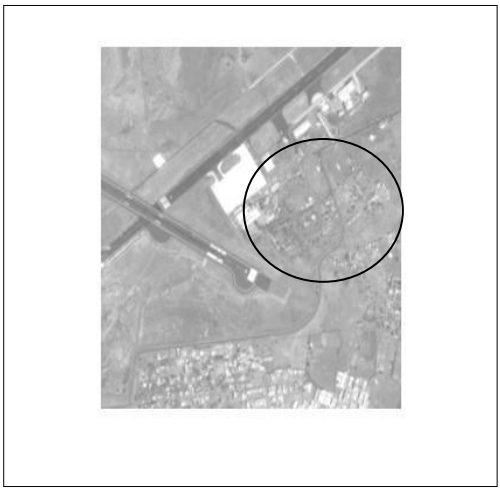
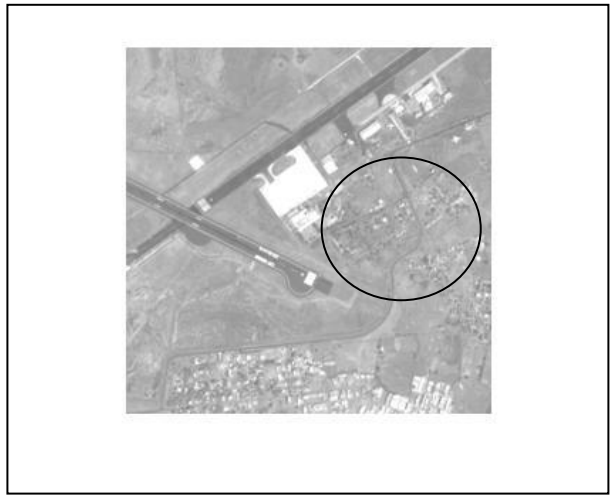

c) Bilinear　　　　　　　　　　　　d) Bicubic

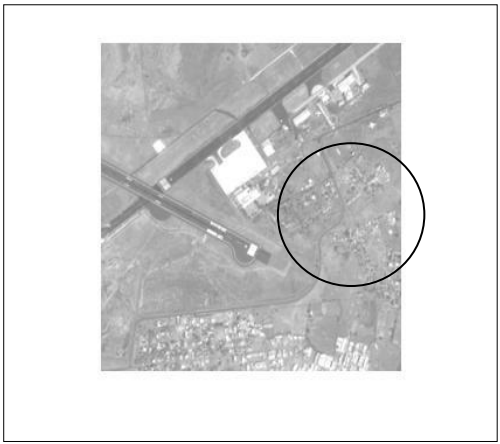

d) CNN based Hybrid Method

**Fig.1** Visual comparison of resampling methods for down sampled Cartosat-1 imagery

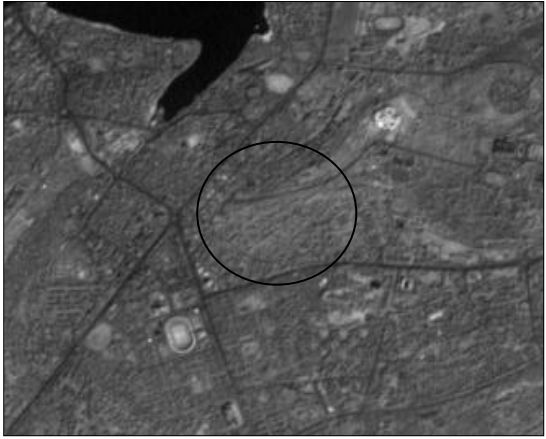

a) PAN sensor image

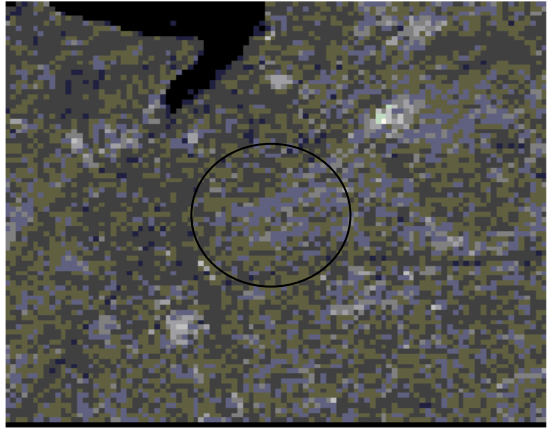

b) Nearest Neighbor

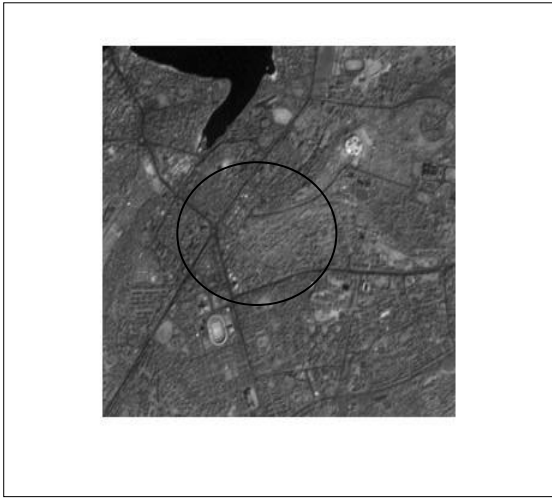

c) Bilinear

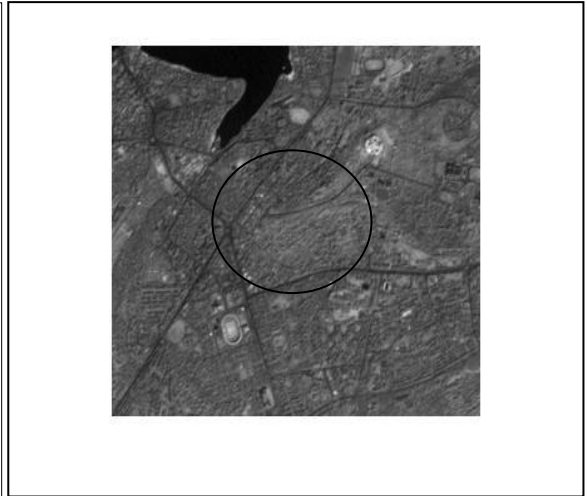

d) Bicubic

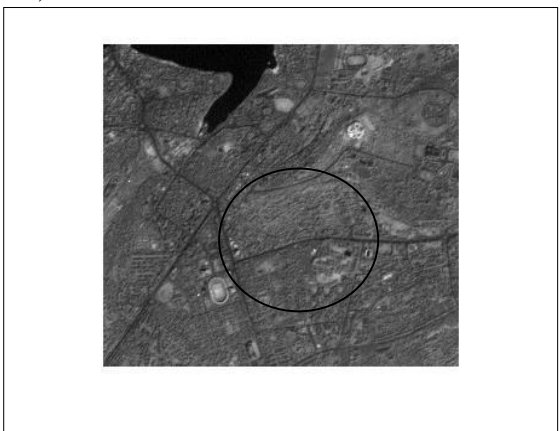

e.) CNN based Hybrid Method

**Fig.2** Visual comparison of different resampling method for down sampled PAN sensor imagery

189                    American Society for Photogrammetry & Remote Sensing

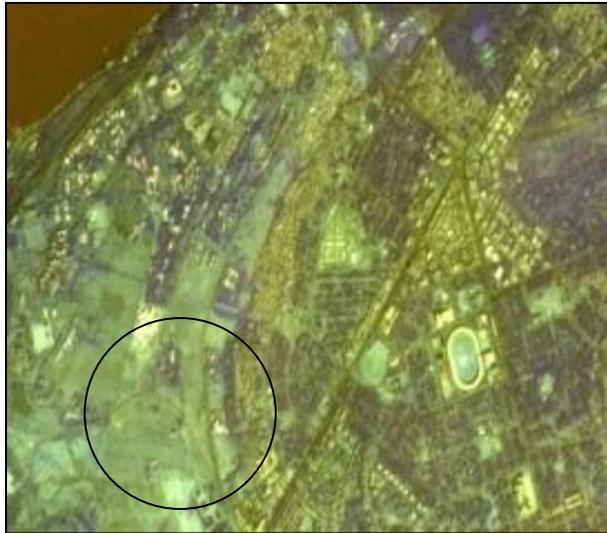
a) LISS 3

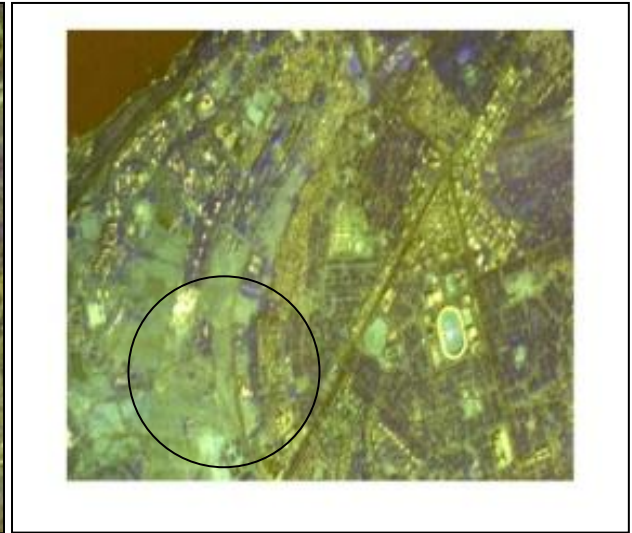
b) Nearest Neighbor

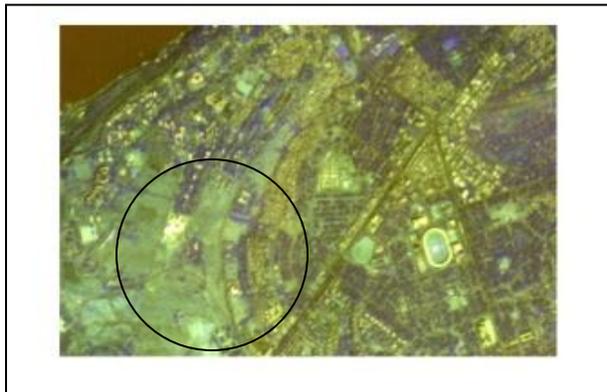
c) Bi Linear

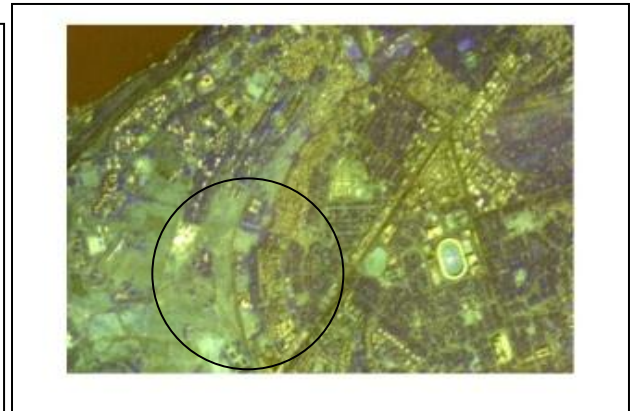
d) Bicubic

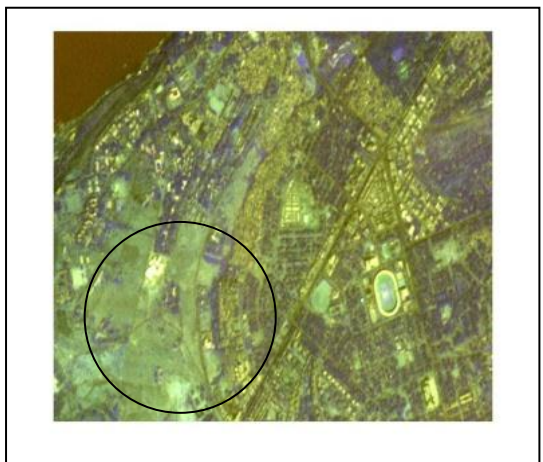
c) CNN Based Method

**Fig.3** Visual comparison of different resampling method for up sampled LISS3 sensor imagery

American Society for Photogrammetry & Remote Sensing

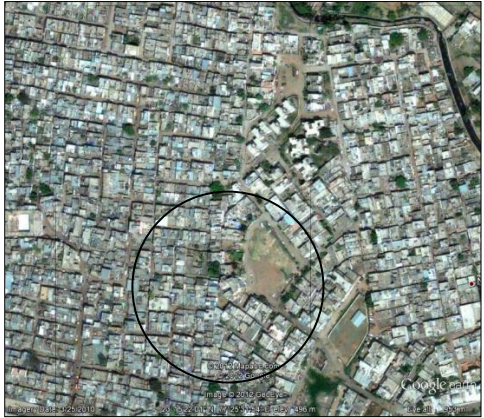
a) Google Earth Imagery

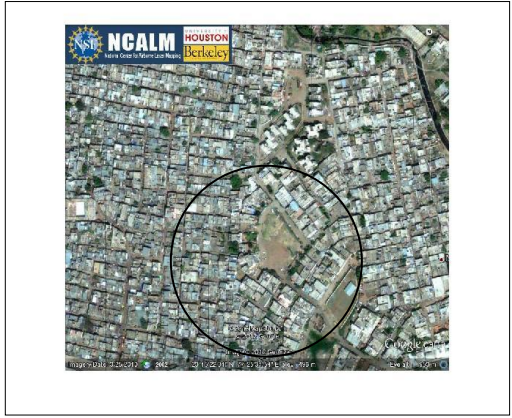
b) Nearest Neighbor

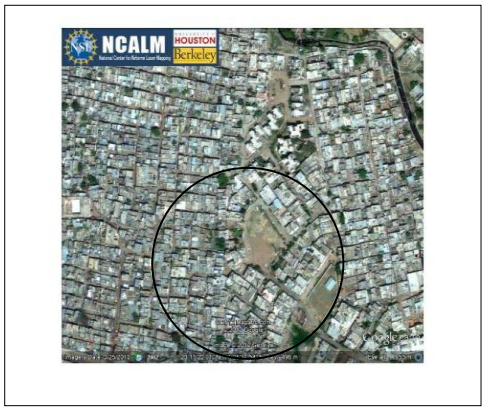
c) Bilinear

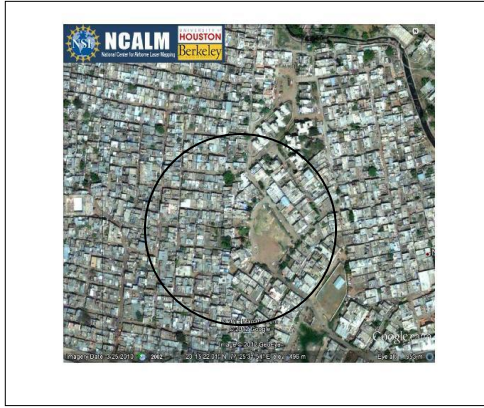
d) Bicubic

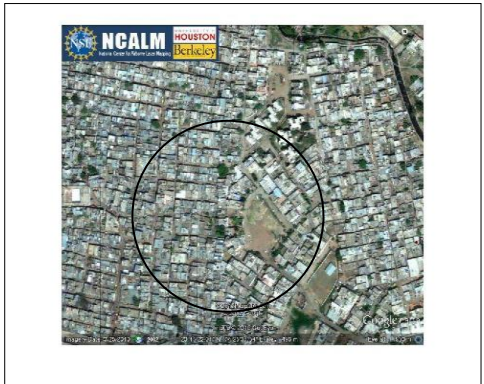
e) CNN Based Hybrid Approach

Fig.4 Visual comparison of resampling methods for up sampled Google Earth imagery

The investigation revealed that the CNN based hybrid approach is advantageous over the existing methodologies for up sampling and down sampling as inferred from the visual results as well as from the statistical measures.

# CONCLUSION

In this paper, we have noted that in a typical image processing and computer vision technique, the simple interpolation step needed to perform registration, may warrant special consideration when working with under sampled images. In particular, we have discussed the fact that naively applying a higher order interpolation technique will most likely degrade the effective sub-pixel detection capabilities. In addition, we have provided a new method that should help in increasing the accuracy of interpolation in the regions of the images which are typically most interesting to human observers: buildings, roads, and other small features which often have sub-pixel fidelity in under sampled data. While the proposed method shows promising results in our early experiments, there is considerable work to be done in precisely characterizing the situations in which it performs optimally.